\documentclass[conference,a4paper]{IEEEtran}
\IEEEoverridecommandlockouts

\usepackage{fancyhdr}

\ifCLASSINFOpdf
\else
\fi

\usepackage{algorithmic}
\usepackage[dvips]{graphicx}
\usepackage{subfigure}
\usepackage{url}
\usepackage{listings}
\usepackage{algorithm}
\usepackage{algorithmic}
\usepackage{multirow}

\newcommand{\bvec}[1]{\mbox{\boldmath $#1$}}

\begin{document}
%
\title{A Video Recognition Method by using Adaptive Structural Learning of Long Short Term Memory based Deep Belief Network
\thanks{\copyright 2019 IEEE. Personal use of this material is permitted. Permission from IEEE must be obtained for all other uses, in any current or future media, including reprinting/republishing this material for advertising or promotional purposes, creating new collective works, for resale or redistribution to servers or lists, or reuse of any copyrighted component of this work in other works.}
}

\author{\IEEEauthorblockN{Shin Kamada}
\IEEEauthorblockA{Advanced Artificial Intelligence Project Research Center,\\
Research Organization of Regional Oriented Studies,\\
Prefectural University of Hiroshima\\
1-1-71, Ujina-Higashi, Minami-ku, \\
Hiroshima 734-8558, Japan\\
E-mail: skamada@pu-hiroshima.ac.jp}
\and
\IEEEauthorblockN{Takumi Ichimura}
\IEEEauthorblockA{Advanced Artificial Intelligence Project Research Center,\\
Research Organization of Regional Oriented Studies,\\
and Faculty of Management and Information System,\\
Prefectural University of Hiroshima\\
1-1-71, Ujina-Higashi, Minami-ku, \\
Hiroshima 734-8558, Japan\\
E-mail: ichimura@pu-hiroshima.ac.jp}
}

\maketitle




\pagestyle{fancy}{
\fancyhf{}
\fancyfoot[R]{}}
\renewcommand{\headrulewidth}{0pt}
\renewcommand{\footrulewidth}{0pt}

\begin{abstract}
  Deep learning builds deep architectures such as multi-layered artificial neural networks to effectively represent multiple features of input patterns. The adaptive structural learning method of Deep Belief Network (DBN) can realize a high classification capability while searching the optimal network structure during the training. The method can find the optimal number of hidden neurons of a Restricted Boltzmann Machine (RBM) by neuron generation-annihilation algorithm to train the given input data, and then it can make a new layer in DBN by the layer generation algorithm to actualize a deep data representation. Moreover, the learning algorithm of Adaptive RBM and Adaptive DBN was extended to the time-series analysis by using the idea of LSTM (Long Short Term Memory). In this paper, our proposed prediction method was applied to Moving MNIST, which is a benchmark data set for video recognition. We challenge to reveal the power of our proposed method in the video recognition research field, since video includes rich source of visual information. Compared with the LSTM model, our method showed higher prediction performance (more than 90\% predication accuracy for test data).  
\end{abstract}


%
\IEEEpeerreviewmaketitle

\section{Introduction}
Recently, Artificial Intelligence (AI) with sophisticated technologies has become an essential technique in our life. \cite{webmarket2016}. Especially, the recent advances in deep learning methods enable higher performance for several big data compared to traditional methods \cite{Bengio09, Quoc12}. For example, CNNs (Convolutional Neural Network) such as AlexNet \cite{AlexNet}, GoogLeNet \cite{GoogLeNet}, VGG16 \cite{VGG16}, and ResNet \cite{ResNet}, highly improved classification or detection accuracy in image recognition \cite{Russakovsky15}. 

As improvement of image recognition, deep learning is also applied to video recognition \cite{Mohammadi18}. The video recognition is kind of fusion task which needs both image recognition and time-series prediction simultaneously. This is, recurrent function that classifies an given image or detects an object while predicting the future, is required. Understanding of time series video is expected in various kinds industrial fields, such as human detection, pose or facial estimation from video camera, autonomous driving system, and so on \cite{Zhang19}.

LSTM (Long Short Term Memory) is a well-known method for time-series prediction and is applied to deep learning methods\cite{Bengio94}. The method enabled the traditional recurrent neural network recognizes not only short-term memory but also long-term memory for given sequential data \cite{Lipton16}. For video recognition of LSTM, the idea using convolutional filter instead of one-dimensional neuron can be used since one frame of sequential video can be seen as one image \cite{Xingjian15}.

In our research, we proposed the adaptive structural learning method of DBN \cite{Kamada18_Springer}. The adaptive structural learning can find a suitable size of network structure for given input space during its training. The neuron generation and annihilation algorithms \cite{Kamada16_SMC, Kamada16_ICONIP} were implemented on Restricted Boltzmann Machine (RBM) \cite{Hinton12}, and layer generation algorithm \cite{Kamada16_TENCON} was implemented on Deep Belief Network (DBN) \cite{Hinton06}. The adaptive structural learning of DBN (Adaptive DBN) shows the highest classification capability in the research field of image recognition by using some benchmark data sets such as MNIST \cite{LeCun98a}, CIFAR-10, and CIFAR-100 \cite{CIFAR10}. Moreover, the learning algorithm of Adaptive RBM and Adaptive DBN was extended to the time-series prediction by using the idea of LSTM \cite{Ichimura17_IJCNN}. LSTM was often implemented on a CNN structure, we implemented LSTM on our Adaptive RBM and DBN, and then the proposed method
showed higher prediction accuracy than the other methods for several time-series benchmark data sets, such as Nottingham (MIDI) and CMU (Motion Capture). 

For further improvement of the method, our proposed method was applied to Moving MNIST \cite{MovingMNIST} in this paper, which is a benchmark data set for video recognition. We challenge to reveal the power of our proposed method in the video recognition research field, since video includes rich source of visual information. Compared with the LSTM model \cite{Srivastava15}, our method make a higher performance of prediction.

The remainder of this paper is organized as follows. In section \ref{sec:adaptive_dbn}, basic idea of the adaptive structural learning of DBN is briefly explained. Section \ref{sec:adaptive_rnndbn} gives the description of the extension algorithm of Adaptive DBN for time-series prediction. In section \ref{sec:exe}, the effectiveness of our proposed method is verified on moving MNIST. In section \ref{sec:conclusion}, we give some discussions to conclude this paper.

\section{Adaptive Learning Method of Deep Belief Network}
\label{sec:adaptive_dbn}
This section explains the traditional RBM \cite{Hinton12} and DBN \cite{Hinton06} to describe the basic behavior of our proposed adaptive learning method of DBN.

\subsection{Neuron Generation and Annihilation Algorithm of RBM}
\label{subsec:adaptive_rbm}
\begin{figure}[tb]
\centering
\includegraphics[scale=0.8]{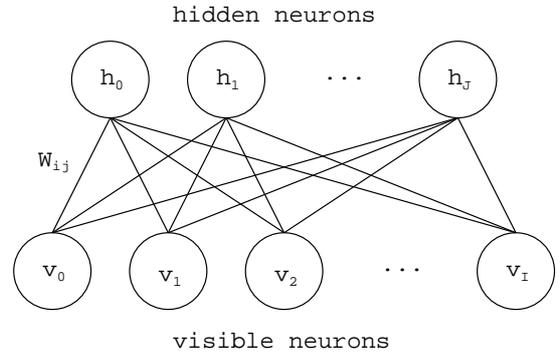}
\caption{A network structure of RBM}
\label{fig:rbm}
\end{figure}

While recent deep learning model has higher classification capability, some problems related to the network structure or the number of some parameters still remains to become a difficult task as the AI research. For the problem, we have developed the adaptive structural learning method in RBM model (Adaptive RBM) \cite{Kamada18_Springer}. RBM as shown in Fig.~\ref{fig:rbm} is an unsupervised graphical and energy based model on two kinds of layers; visible layer for input and hidden layer for feature vector, respectively. The neuron generation algorithm of the Adaptive RBM can generate an optimal number of hidden neurons and the trained RBM is suitable structure for given input space.

\begin{figure}[tbp]
\begin{center}
\subfigure[Neuron generation]{\includegraphics[scale=0.5]{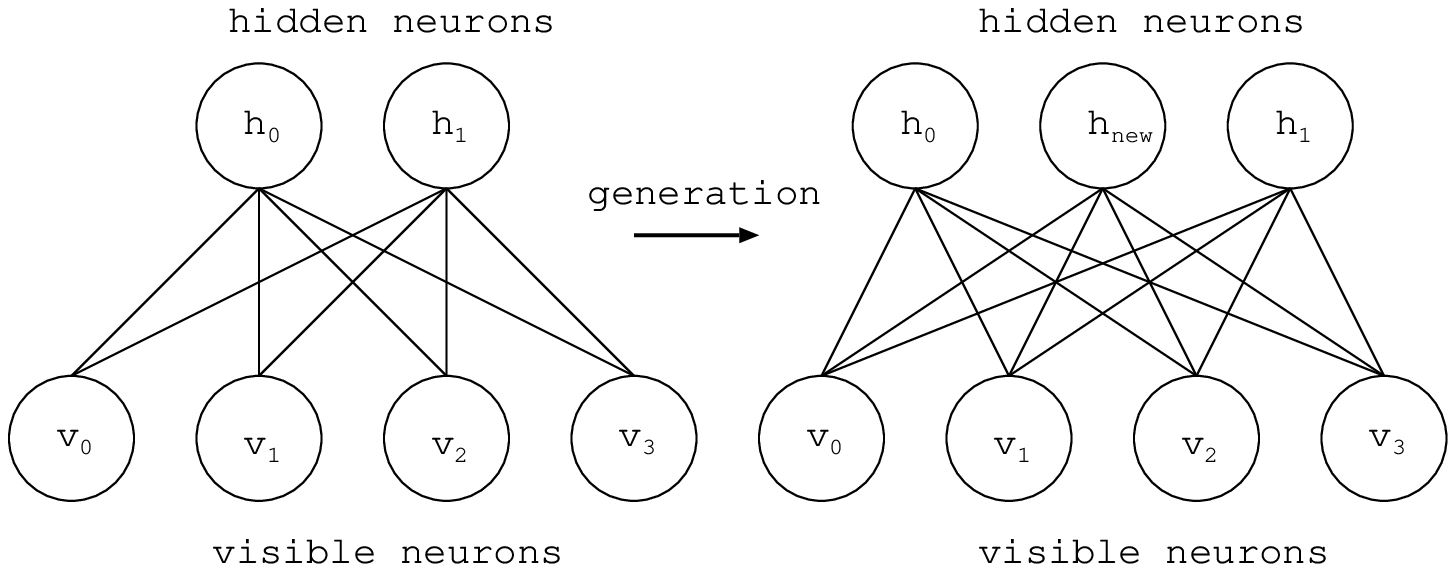}\label{fig:neuron_generation}}
\subfigure[Neuron annihilation]{\includegraphics[scale=0.5]{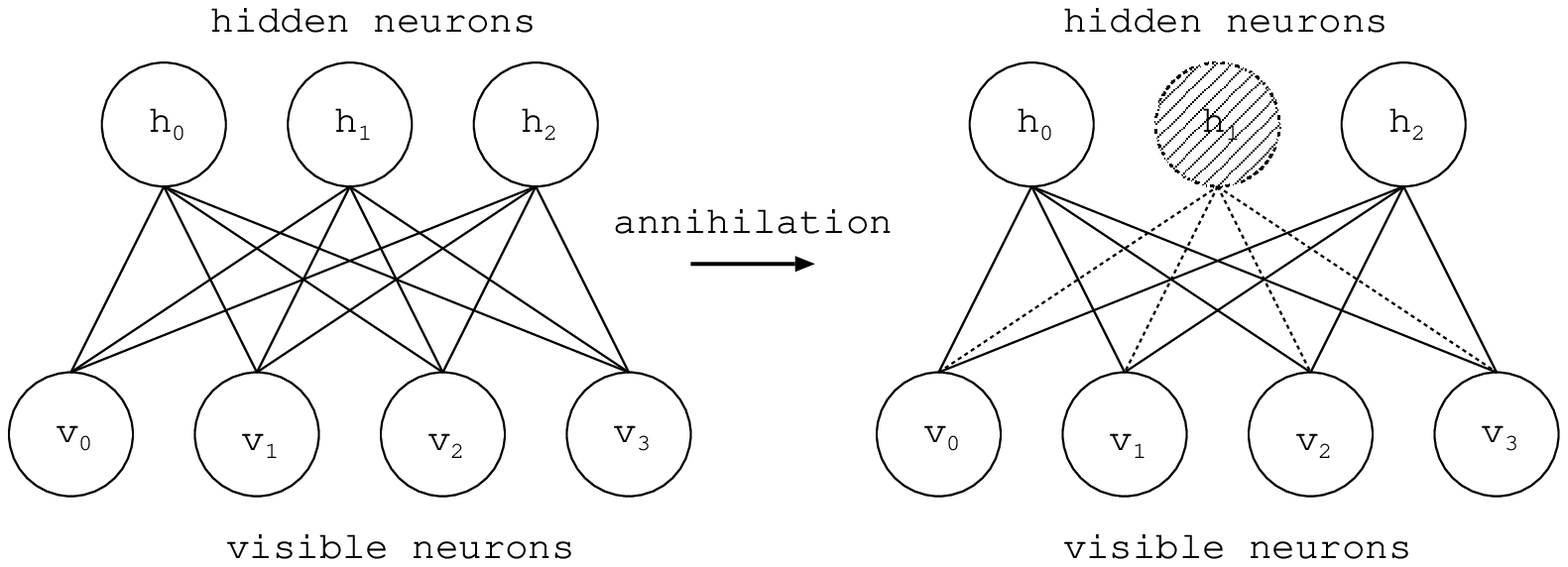}\label{fig:neuron_annihilation}}
\caption{Adaptive RBM}
\label{fig:adaptive_rbm}
\end{center}
\end{figure}

The neuron generation is based on the idea of Walking Distance (WD), which is inspired from the multi-layered neural network in the paper \cite{Ichimura95}. WD is the difference between the prior variance and the current one of learning parameters.
RBM has 3 kinds of parameters according to visible neurons, hidden neurons, and the weights among their connections. The Adaptive RBM can monitor their parameters excluding the visible one ( The paper \cite{Kamada18_Springer} describes the reason of the disregard). The situation means that only the existing hidden neurons cannot represent an ambiguous pattern, because there is the lack of the number of hidden neurons. In order to express the ambiguous patterns, a new neuron is inserted to inherit the attributes of the parent hidden neuron as shown in Fig.~\ref{fig:neuron_generation}.

In addition to the neuron generation, the neuron annihilation algorithm was applied to the Adaptive RBM after neuron generation process as shown in Fig.~\ref{fig:neuron_annihilation}. We may meet that some unnecessary or redundant neurons were generated due to the neuron generation process. Therefore, such neurons will be removed the corresponding hidden neuron according to the output activities.

\subsection{Layer Generation Algorithm of DBN}
\label{subsec:adaptive_dbn}
A DBN is a hierarchical model of stacking the several pre-trained RBMs. For building process, output (hidden neurons activation) of $l$-th RBM can be seen as the next input of $(l+1)$-th RBM. Generally, DBN with multiple RBMs has higher data representation power than one RBM. Such hierarchical model can represent the specified features from an abstract concept to an concrete object in the direction from input layer to output layer. However, the optimal number of RBMs depends on the target data space.

\begin{figure*}[tbp]
\centering
\includegraphics[scale=1.0]{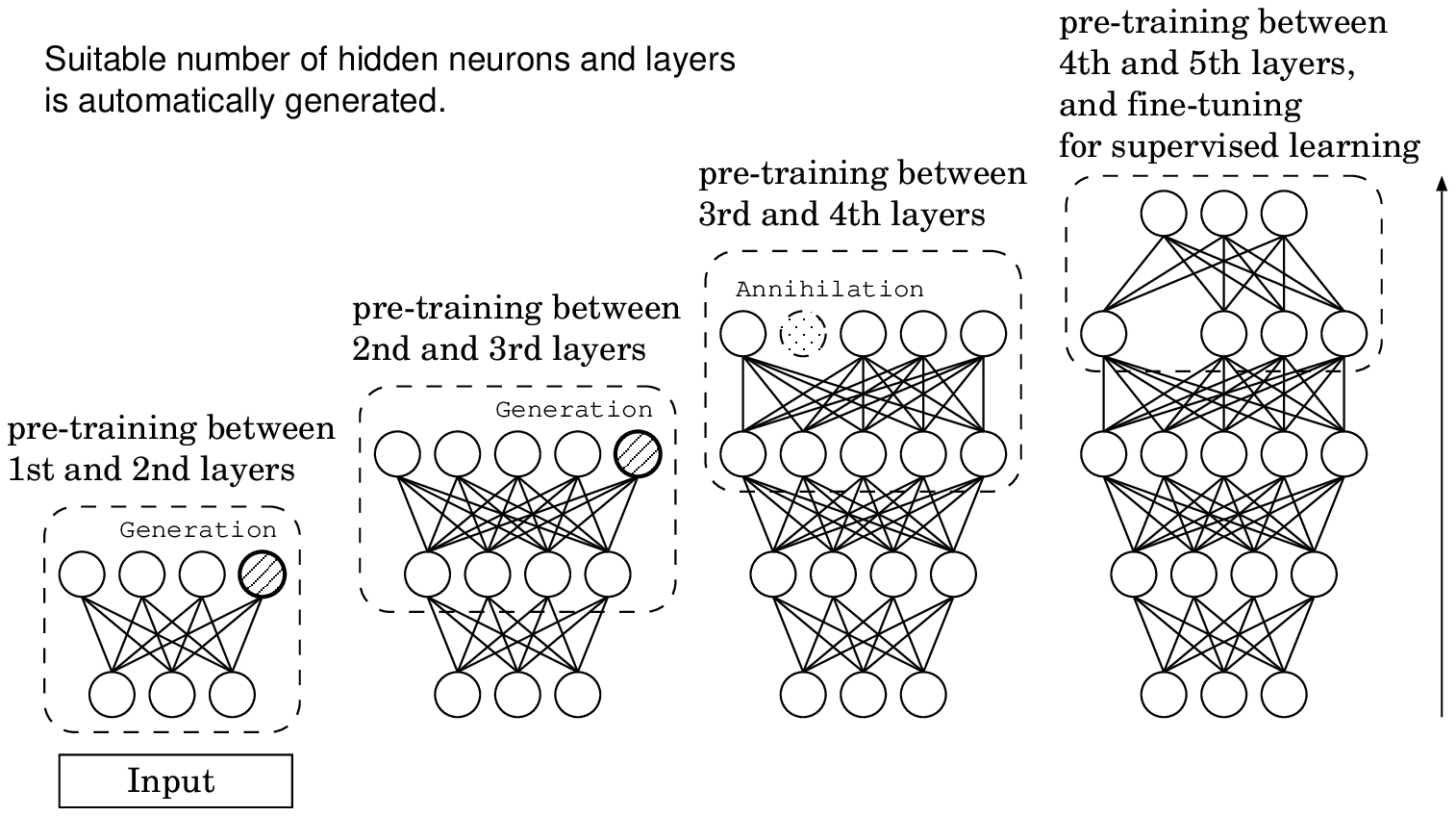}
\caption{Overview of Adaptive DBN}
\label{fig:adaptive_dbn}
\end{figure*}

We developed Adaptive DBN which can automatically adjust an optimal network structure by the self-organization in the similar way of WD monitoring. If both WD and the energy function do not become small values, then a new RBM will be generated to keep the suitable network classification power for the data set, since the RBM has lacked the power of data representation to draw an image of input patterns. Therefore, the condition for layer generation is defined by using the total WD and the energy function. Fig.~\ref{fig:adaptive_dbn} shows the overview of layer generation in Adaptive DBN.

\section{Adaptive RNN-DBN for time-series prediction}
\label{sec:adaptive_rnndbn}
In time-series prediction, some LSTM methods improve the prediction performance of the traditional recurrent neural network by using the several gates such as forget-gate, peephole connection gate, full and gradient gate \cite{Bengio94}. These gates can represent multiple patterns of time-series sequence, that is not only short-term memory but also long-term memory.

Recurrent Neural Network Restricted Boltzmann Machine (RNN-RBM) \cite{Lewandowski12} is a RBM based recurrent model for time-series prediction. The method is a extension model of the traditional Temporal RBM (TRBM) and Recurrent TRBM (RTRBM) \cite{Sutskever08} and it also used a similar idea of LSTM for better performance. 

\begin{figure}[tbp]
\begin{center}
\includegraphics[scale=0.55]{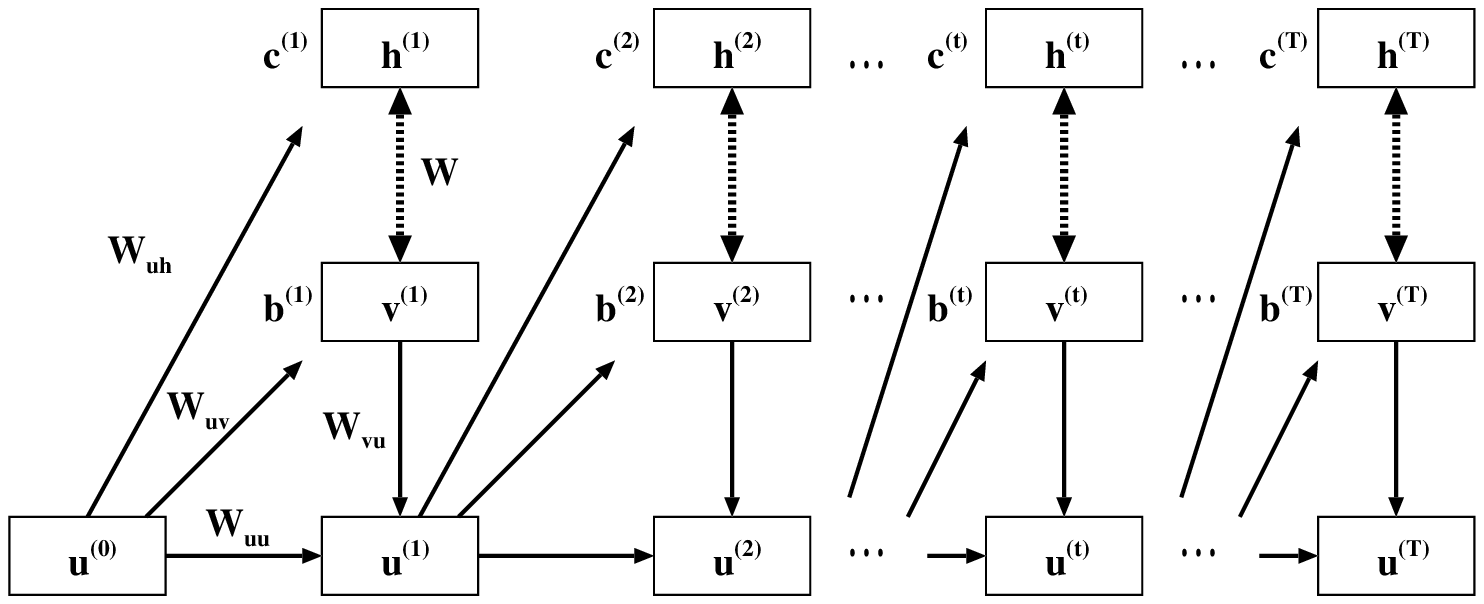}
\caption{Structure of RNN-RBM}
\label{fig:rnn-rbm}
\end{center}
\end{figure}

Fig.~\ref{fig:rnn-rbm} shows the network structure of RNN-RBM. RNN-RBM had a recurrent structure on Markov process for given time-series sequence, as well as RNN. Let an input sequence with the length $T$ be $\bvec{V} = \{ \bvec{v}^{(1)}, \cdots, \bvec{v}^{(t)}, \cdots, \bvec{v}^{(T)} \}$. $\bvec{v}^{(t)}$ and $\bvec{h}^{(t)}$ are the input and hidden neurons at $t$-th RBM, respectively. The time dependency parameters $\bvec{b}^{(t)}$, $\bvec{c}^{(t)}$, and $\bvec{u}^{(t)}$ are calculated by the past parameters at $t-1$ as following equations. 
\begin{equation}
\label{eq:calc_bt}
\bvec{b}^{(t)} = \bvec{b} + \bvec{W}_{uv} \bvec{u}^{(t-1)},
\end{equation}
\begin{equation}
\label{eq:calc_ct}
\bvec{c}^{(t)} = \bvec{c} + \bvec{W}_{uh} \bvec{u}^{(t-1)},
\end{equation}
\begin{equation}
\label{eq:calc_ut}
\bvec{u}^{(t)} = \sigma( \bvec{u} + \bvec{W}_{uu} \bvec{u}^{(t-1)} + \bvec{W}_{vu} \bvec{v}^{(t)} ),
\end{equation}
where $\sigma()$ is a activation function. For example, $Tanh$ function was used in the paper \cite{Lewandowski12}. $\bvec{u}^{(t)}$ represents time-series context for give input. $\bvec{{u}}^{(0)}$ is the initial state and the values are given randomly. $\bvec{\theta}=\{\bvec{b}, \bvec{c}, \bvec{W}, \bvec{u}, \bvec{W}_{uv}, \bvec{W}_{uh}, \bvec{W}_{vu}, \bvec{W}_{uu} \}$ is a set of learning parameters, and is not time-dependency. At each time $t$, the RBM learning can employ the update algorithm of $\bvec{b}^{(t)}$ and $\bvec{c}^{(t)}$ at time $t$, and weights $\bvec{W}$ between them. After the calculation of error continues till time $T$, the gradient for $\bvec{\theta}$ is updated to trace from time $T$ back to time $t$ to the contrary by BPTT\cite{Elman90,Jordan86}. BPTT is Back Propagation Through Time method which is often used in the traditional recurrent neural network.

We developed Adaptive RNN-RBM by applying the neuron generation and annihilation algorithm to RNN-RBM. An suitable number of hidden neurons is sought by the neuron generation and annihilation algorithm for better representation power as same as usual Adaptive RBM \cite{Ichimura17_IJCNN}. In addition, the hierarchical model with layer generation was developed as Adaptive RNN-DBN by using the same monitoring function of Adaptive DBN. In other words, the output signal of hidden neuron $\bvec{h}^{(t)}$ at $l$-th Adaptive RNN-RBM can be seen as the next input signal at $(l+1)$-th Adaptive RNN-RBM as shown in Fig.~\ref{fig:adaptive_dbn} The proposed Adaptive RNN-DBN was applied to several time-series benchmark data sets, such as Nottingham (MIDI) and CMU (Motion Capture). In \cite{Ichimura17_IJCNN}, the prediction accuracy of the proposed method is higher than that of the traditional methods.

\section{Experiment Results}
\label{sec:exe}
In this paper, the effectiveness of our proposed method for moving MNIST benchmark data set \cite{Srivastava15} was verified. We challenge to reveal the power of our proposed method in the video recognition research field, since video includes rich source of visual information. When we want to detect and identify visual objects, faces, emotions, actions or events in the real time, state-of-the-art video recognition software for the comprehensive video content analysis can make the advanced solutions for AI-based visual content search. Therefore, the prediction performance is compared with the traditional LSTM in this paper.

\begin{figure*}[tbp]
\begin{center}
\includegraphics[scale=0.65]{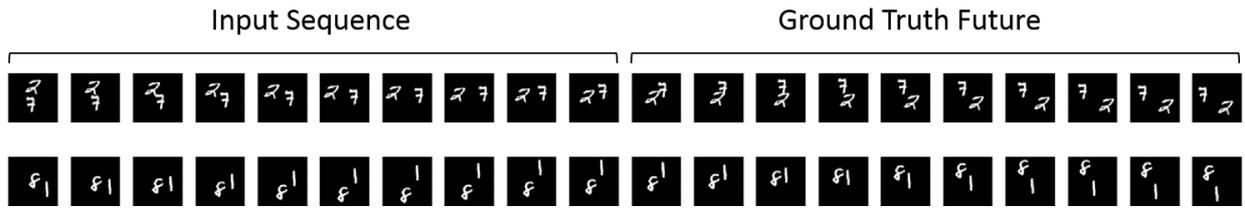}
\caption{Two samples of Moving MNIST}
\label{fig:moving_mnist_sample}
\end{center}
\end{figure*}

\subsection{Data set: Moving MNIST}
Moving MNIST \cite{Srivastava15} is a benchmark data set for video recognition. There are 10,000 samples including 8,000 for training and 2,000 for test. Each sample consists of 20 sequential gray scale images ($[0, 255]$) of $64 \times 64$ patch, where two digits move in the sequences. The digits were chosen randomly from the training set. The selected digits are placed initially at random locations inside the patch. Each digit was assigned a velocity which has direction and magnitude. The direction and the magnitude also chosen randomly. Fig.~\ref{fig:moving_mnist_sample} shows two samples of 20 sequential images.

Moving MNIST is used in various researches such as the video decomposition \cite{Hsieh18} and the future predictor \cite{Srivastava15}. This paper aims to investigate the effectiveness of the our LSTM model in the video recognition and we compare the recognition capability as future predictor.

\begin{figure}[tbp]
\begin{center}
\includegraphics[scale=0.65]{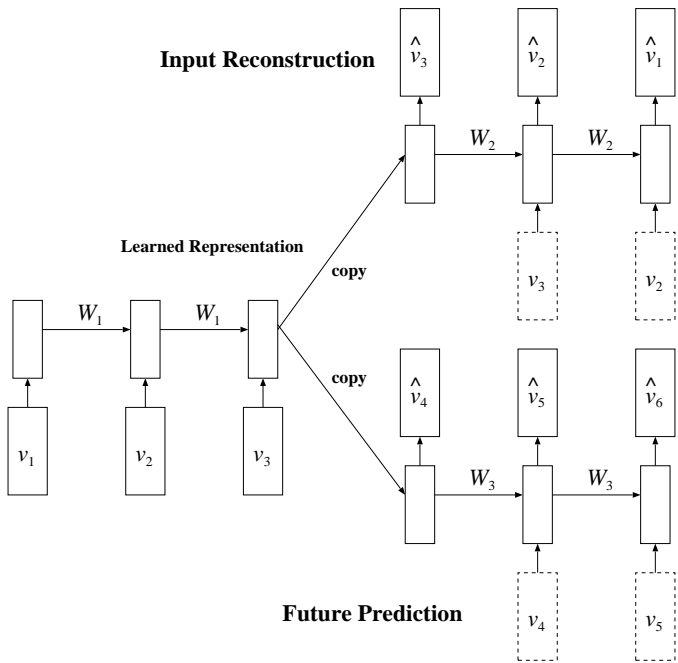}
\caption{The Composite Model of LSTM \cite{Srivastava15}}
\label{fig:lstm_composite_model}
\end{center}
\end{figure}

Since any teacher signal for each sample is not provided in moving MNIST, the LSTM in \cite{Srivastava15} investigated the cross entropy between the given sequence (ground truth) and the predicted sequence. Fig.~\ref{fig:lstm_composite_model} shows the composite model of LSTM in \cite{Srivastava15}. In the method, the first 10 frames are used as an input sequence of the model, the remaining 10 frames are used for evaluation of future prediction as shown in Fig.~\ref{fig:moving_mnist_sample}.

\begin{figure*}[tbp]
\begin{center}
\includegraphics[scale=0.6]{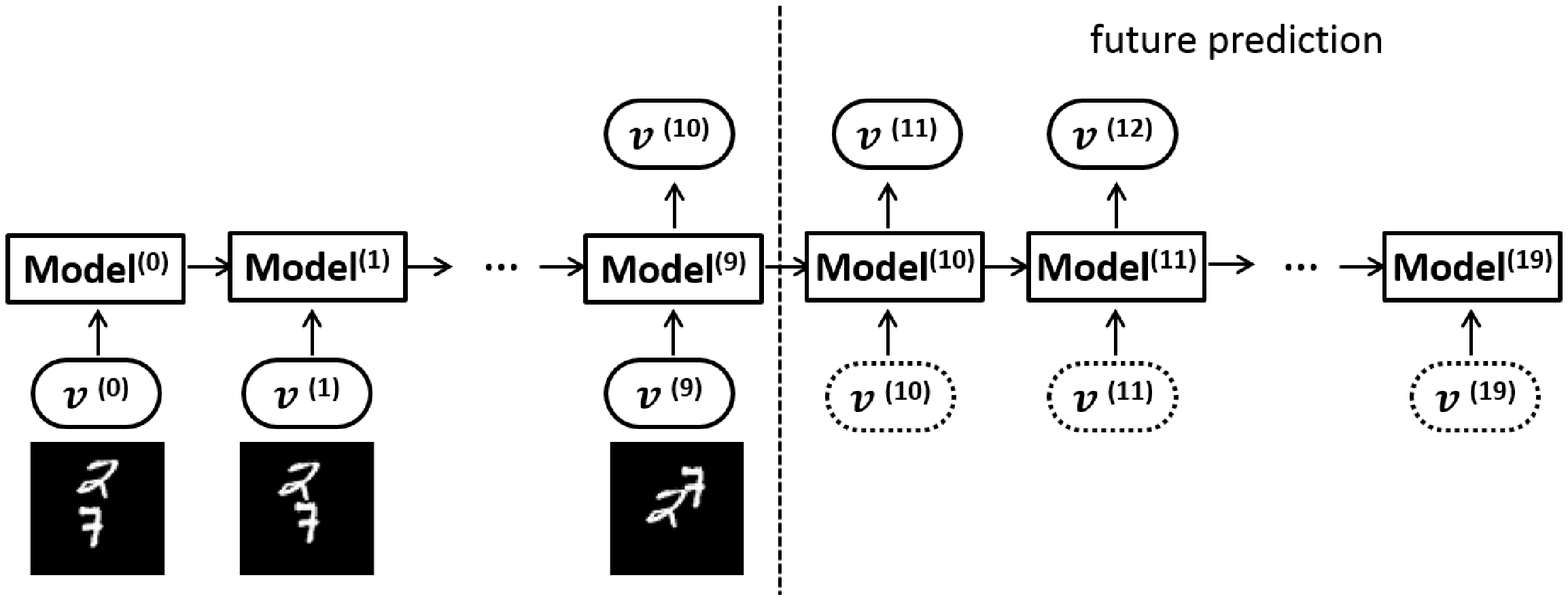}
\caption{Abstract of prediction}
\label{fig:prediction_abstract}
\end{center}
\end{figure*}

Fig.~\ref{fig:prediction_abstract} shows an abstract procedure of prediction in our method. As same as in our method, same evaluation method was used. Since our method can predict next frame for given frame, the predicted frame is used for next input, and then squared error and prediction accuracy of the ground truth images and the predicted images were evaluated.

\subsection{Experimental Results}
Table \ref{tab:result1} and Table \ref{tab:result2} show the prediction result on moving MNIST for training and test. The value in Table \ref{tab:result1} is the cross entropy between the ground truth images and the predicted images at last sequence. The result of the LSTM is cited from the paper \cite{Srivastava15}. The left value of each column in Table \ref{tab:result2} is the squared error and the value in brackets is the prediction accuracy. 

In the LSTM, the result for training and the prediction accuracy for test are not reported. In the Adaptive RNN-DBN, three settings of learning ratio (lr) were used for evaluation. For training of the Adaptive RNN-DBN, the cross entropy and the squared error reached almost 0 and prediction accuracy reached almost 100\%. For test, the three settings of Adaptive RNN-DBN showed smaller prediction error than the LSTM. The best performance was acquired when the learning ratio was 0.050 in the Adaptive RNN-DBN.

We also investigated the intermediate result in the predicted images of the Adaptive RNN-DBN. Table \ref{tab:result3} expresses the future prediction accuracy for each predicted sequence. Basically, the prediction accuracy was slowly decreased from 11th to 20th frames.

\begin{table}[b]
\caption{Prediction result (Cross entropy)}
\vspace{-5mm}
\label{tab:result1}
\begin{center}
\begin{tabular}{c|c|c}
\hline 
Model & Training & Test  \\
\hline
LSTM \cite{Srivastava15}  & - & 341.2 \\ 
Adaptive RNN-DBN (lr = 0.010) & 18.5  & 165.3 \\
Adaptive RNN-DBN (lr = 0.050) & 16.1  & 134.0 \\ 
Adaptive RNN-DBN (lr = 0.001) & 17.0  & 140.6 \\ 
\hline
\end{tabular}
\end{center}
\end{table}

\begin{table}[b]
\caption{Prediction result (Squared loss and correct ratio)}
\vspace{-5mm}
\label{tab:result2}
\begin{center}
\begin{tabular}{c|cc|cc}
\hline 
Model & \multicolumn{2}{c|}{Training} &  \multicolumn{2}{c}{Test}  \\
\hline
Adaptive RNN-DBN (lr = 0.010) & 11.4 & (99.0\%)  & 119.9 & (91.8\%) \\
Adaptive RNN-DBN (lr = 0.050) & 10.3 & (99.4\%)  & 100.5 & (92.5\%) \\ 
Adaptive RNN-DBN (lr = 0.001) & 14.5 & (98.9\%)  & 140.2 & (89.4\%) \\ 
\hline
\end{tabular}
\end{center}
\end{table}

\begin{table*}[tbp]
\caption{Prediction Accuracy for each frame}
\label{tab:result3}
\begin{center}
\begin{tabular}{c|c|c|c|c|c|c|c|c|c|c}
  \hline
& \multicolumn{10}{c}{Sequence} \\ \cline{2-11} 
Model & 11  & 12  & 13 & 14 & 15 & 16 & 17 & 18 & 19 & 20  \\ 
\hline
Adaptive RNN-DBN (lr = 0.010) & 98.2\%  & 97.9\% & 97.0\% & 95.2\% & 94.5\% & 93.9\% & 93.3\% & 92.8\% & 92.0\% & 91.8\%   \\
Adaptive RNN-DBN (lr = 0.050) & 99.5\%  & 98.0\% & 97.9\% & 96.5\% & 96.4\% & 94.5\% & 93.1\% & 92.9\% & 92.8\% & 92.5\%   \\ 
Adaptive RNN-DBN (lr = 0.001) & 96.8\%  & 96.8\% & 96.4\% & 94.8\% & 93.6\% & 92.9\% & 91.8\% & 91.0\% & 90.0\% & 89.4\%   \\ 
\hline
\end{tabular}
\end{center}
\end{table*}

\section{Conclusion}
\label{sec:conclusion}
Deep learning is widely used in various kinds of research fields, especially image recognition. In our research, Adaptive DBN which can find the optimal network structure for given data was developed. The method shows higher classification accuracy than existing deep learning methods for several benchmark data sets. In this paper, our proposed prediction method was applied to Moving MNIST, which is a benchmark data set for video recognition. Compared with the LSTM model, our method showed higher prediction performance (more than 90\% predication accuracy for test data). Our proposed method will be further improved for better prediction capability by evaluating the method on the other large video databases such as video streaming and defect detection in time-series video data.

\section*{Acknowledgment}
This work was supported by JSPS KAKENHI Grant Number 19K12142, 19K24365, and obtained from the commissioned research by National Institute of Information and Communications Technology (NICT, 21405), JAPAN.



%

\end{document}